%% file: iclr2027_conference.tex
\documentclass{article} 
\usepackage{iclr2027_conference,times}

\input{math_commands.tex}

\usepackage{hyperref}
\usepackage{url}
\usepackage{booktabs}
\usepackage{multirow}
\usepackage{graphicx}
\usepackage{amsmath,amssymb}
\usepackage{flafter}
\usepackage{graphicx}
\usepackage{booktabs}
\usepackage{capt-of}
\usepackage{needspace}
\usepackage{wrapfig}
\title{CREDO: Evidence-Conditioned Credit Assignment for Temporal Reasoning Agents}

\author{
\textbf{%
Yue Leng\textsuperscript{1}\thanks{Equal contribution.},
Shengtao Wen\textsuperscript{1}\footnotemark[1],
Xiang Chen\textsuperscript{1}\thanks{Corresponding author.},
Xinyi Zeng\textsuperscript{2},
Yu Tian\textsuperscript{3},
Sheng-Jun Huang\textsuperscript{1}\footnotemark[2]
}\\
{\normalfont
\textsuperscript{1}\,Nanjing University of Aeronautics and Astronautics
}\\
{\normalfont
\textsuperscript{2}\,Tsinghua University
}\\
{\normalfont
\textsuperscript{3}\,360 Security Capability Center, Qihoo 360
}\\
{\normalfont
\texttt{\{lengyue,xiang\_chen\}@nuaa.edu.cn}
}
}

\iclrfinalcopy

\begin{document}

\maketitle

\fancyhead{}
\renewcommand{\headrulewidth}{0pt}
\fancyfoot[C]{\thepage}

\hypersetup{hidelinks}

\newcommand{\bestscore}[1]{%
  \begingroup
  \setlength{\fboxsep}{1.5pt}%
  \colorbox[HTML]{E2F2F3}{\textbf{#1}}%
  \endgroup
}

\newcommand{\secondscore}[1]{%
  \begingroup
  \setlength{\fboxsep}{1.5pt}%
  \colorbox[HTML]{FCE9ED}{\underline{#1}}%
  \endgroup
}

\newcounter{credoalgorithm}
\newcommand{\credoalgorithmname}{Algorithm}

\makeatletter
\providecommand{\CREDOclearwrap}{\par\WFclear}
\makeatother

\graphicspath{{figures/}}
\setlength{\columnsep}{12pt}
\raggedbottom

\fancyhead{}
\renewcommand{\headrulewidth}{0pt}

\suppressfloats[t]
\suppressfloats[b]

\begin{abstract}
Temporal Knowledge Graph Question Answering (TKGQA) requires language model agents to retrieve evidence, select relevant facts, and perform multi-step temporal reasoning. However, the trajectory-level outcome signal in GRPO does not explicitly distinguish the quality of intermediate decisions: downstream failures can penalize useful evidence acquisition, while identical rewards in all-failed groups eliminate the relative outcome signal despite differences in intermediate progress. We propose Credit Redistribution with Evidence-Driven Opportunities (CREDO), which uses intermediate evidence to recover local supervision, including in all-failed groups, and determine where outcome supervision acts. CREDO assigns positive stage-level credit through comparisons among rollouts of the same question and uses retained answer overlap as an opportunity proxy to shift part of the outcome-policy weight toward reasoning and answering. The framework preserves terminal rewards and trajectory-level advantages, while redistribution conserves total outcome-policy weight per trajectory. Following the public evaluation protocol of prior work, CREDO raises Overall accuracy from 78.0\% to 82.4\% and Multiple accuracy from 55.0\% to 65.6\% on MultiTQ relative to the strongest published baseline in our comparison. On TimelineKGQA, it raises cross-KG accuracy from 30.9\% to 36.4\% relative to our reproduced baseline via direct transfer to ICEWS Actor under matched training and evaluation settings
\footnote{Code is available at
\url{https://github.com/NUAA-MMMI/CREDO}}.
\end{abstract}

\section{Introduction}

\begin{figure}[!t]
\centering
\includegraphics[width=\linewidth]{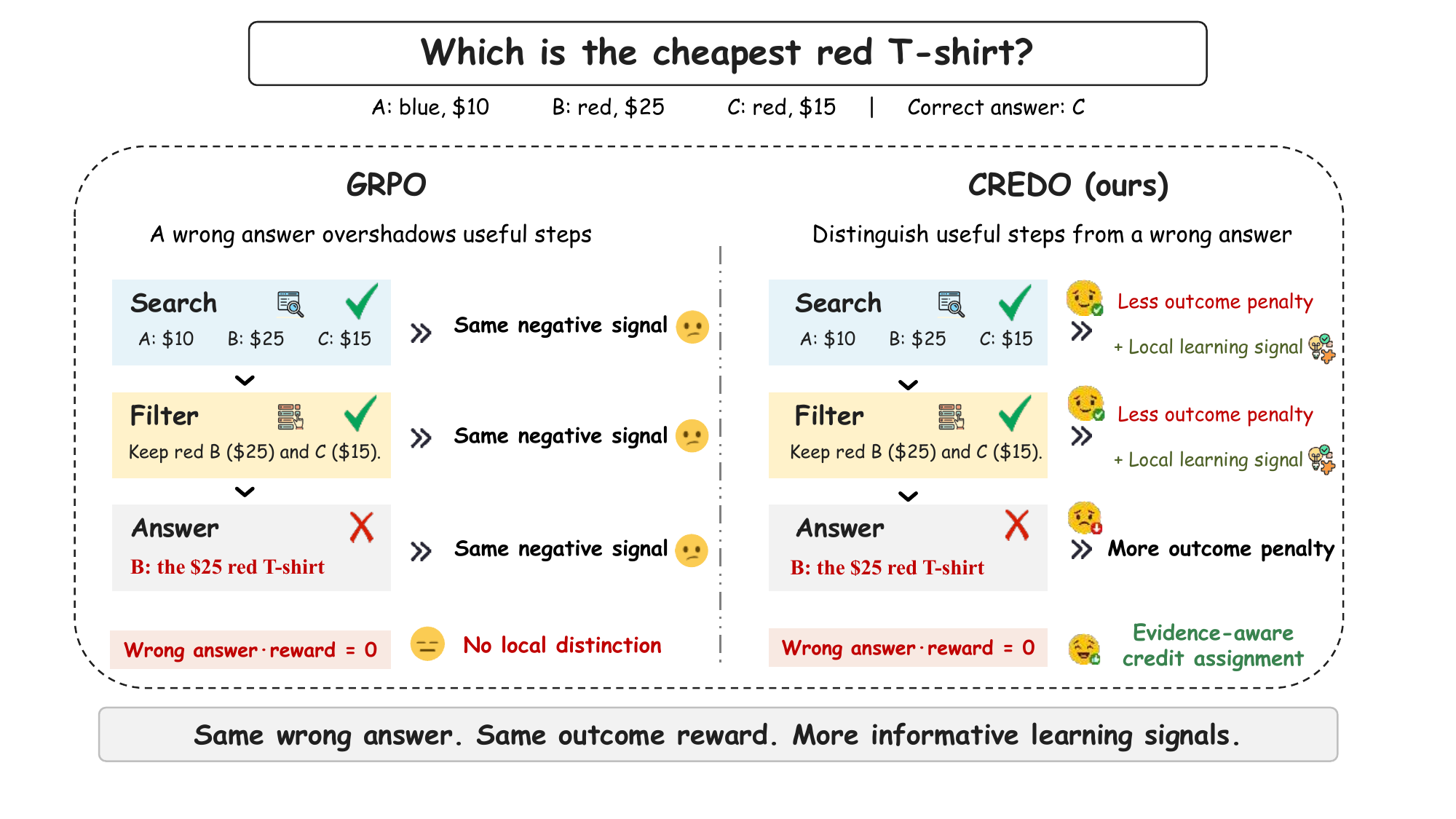}
\caption{\textbf{Motivation of CREDO.} In a mixed-outcome group, useful intermediate decisions can receive negative outcome credit when the final answer is wrong. CREDO uses intermediate evidence for within-question comparisons and evidence-conditioned outcome redistribution.}
\label{fig:credo-motivation}
\end{figure}

Temporal Knowledge Graph Question Answering (TKGQA) requires reasoning over time-dependent facts under temporal constraints \citep{saxena2021temporal,chen2023multitq}. Recent advances in large language models have improved temporal reasoning through structured retrieval, evidence construction, and experience reuse \citep{hu2026sar,liu2026tecqa,tan2026memotime}. Agent-based approaches further integrate retrieval and reasoning into dynamic interaction trajectories, allowing models to adapt decisions to new evidence \citep{qianyihu-etal-2025-TempAgent,lv2026agentsearch,jiang2026tkgthinker,huang2026temptool}. However, an agent may retrieve useful facts but fail to retain or reason over them correctly. The final answer therefore \emph{only partially reflects intermediate decision quality}, motivating closer examination of how temporal agents learn from their interaction trajectories.

Reinforcement learning enables language models to learn multi-step reasoning and search strategies from outcome supervision \citep{shao2024deepseekmath,guo2025deepseek-R1,jin2025searchr1,song2025r1searcher,chen2025research}, with recent extensions to temporal agents \citep{gong2026tempr1,huang2026temptool}. However, outcome-supervised GRPO assigns the \emph{same terminal-reward-based advantage} to all trainable tokens within a trajectory, without explicitly distinguishing intermediate decision quality \citep{shao2024deepseekmath}. Process supervision and fine-grained credit assignment provide more localized learning signals \citep{lightman2024verify,wang2024mathshepherd,xie2026tips,zong2026at2po,zhang2026criticsearch}. For temporal agents, credit assignment must also account for decision dependencies: evidence selection relies on what retrieval exposes, while answer derivation relies on what selection retains. This raises a central question: how can intermediate evidence guide both what agents learn from unsuccessful trajectories and where terminal supervision should act?

\noindent \textbf{Failed trajectories can still contain useful evidence.} An agent may retrieve and retain relevant temporal facts yet answer incorrectly (Figure~\ref{fig:credo-motivation}). In mixed-outcome groups, useful retrieval and selection decisions can receive negative outcome credit. When all rollouts fail with identical rewards, outcome advantages vanish, leaving differences in intermediate evidence quality unexploited. These cases motivate distinguishing \emph{local behavior quality} from \emph{evidence opportunity}: how well a stage performs and how much answer-related evidence upstream decisions provide.

\noindent\textbf{Credit redistribution with evidence-driven opportunities.} Motivated by this analysis, we propose Credit Redistribution with Evidence-Driven Opportunities (CREDO), which uses intermediate evidence to guide both local supervision and outcome credit. CREDO compares retrieval coverage, evidence retention, and answer quality across rollouts of the same question, \emph{rewarding locally superior behavior while conditioning downstream credit on upstream evidence}. These comparisons can recover local learning signals \emph{even in all-failed groups}. For eligible trajectories with retained answer-related evidence, CREDO also shifts a controlled fraction of outcome-policy weight toward reasoning and answering tokens. This redistribution applies to both positive and negative advantages while \emph{preserving total outcome-policy weight per trajectory}. Both components retain the original terminal reward and trajectory-level GRPO advantage without requiring an additional learned evaluator. Answer overlap with retrieved and retained facts serves as a proxy for evidence opportunity.

\noindent \textbf{Contributions.} Our main contributions are threefold. (1) We analyze credit assignment in temporal agents through dependencies among retrieval, evidence selection, and answering, distinguishing local behavior quality from upstream evidence opportunity when learning from failed trajectories. (2) We propose CREDO, which distinguishes local quality from upstream evidence opportunity to recover stage-level supervision through within-question comparisons, including in all-failed groups. (3) We develop evidence-conditioned outcome redistribution that shifts policy weight toward reasoning and answering for both positive and negative trajectory advantages while preserving total outcome-policy weight per trajectory, terminal rewards, and GRPO advantages.

\begin{figure}[t]
\centering
\includegraphics[width=\linewidth]{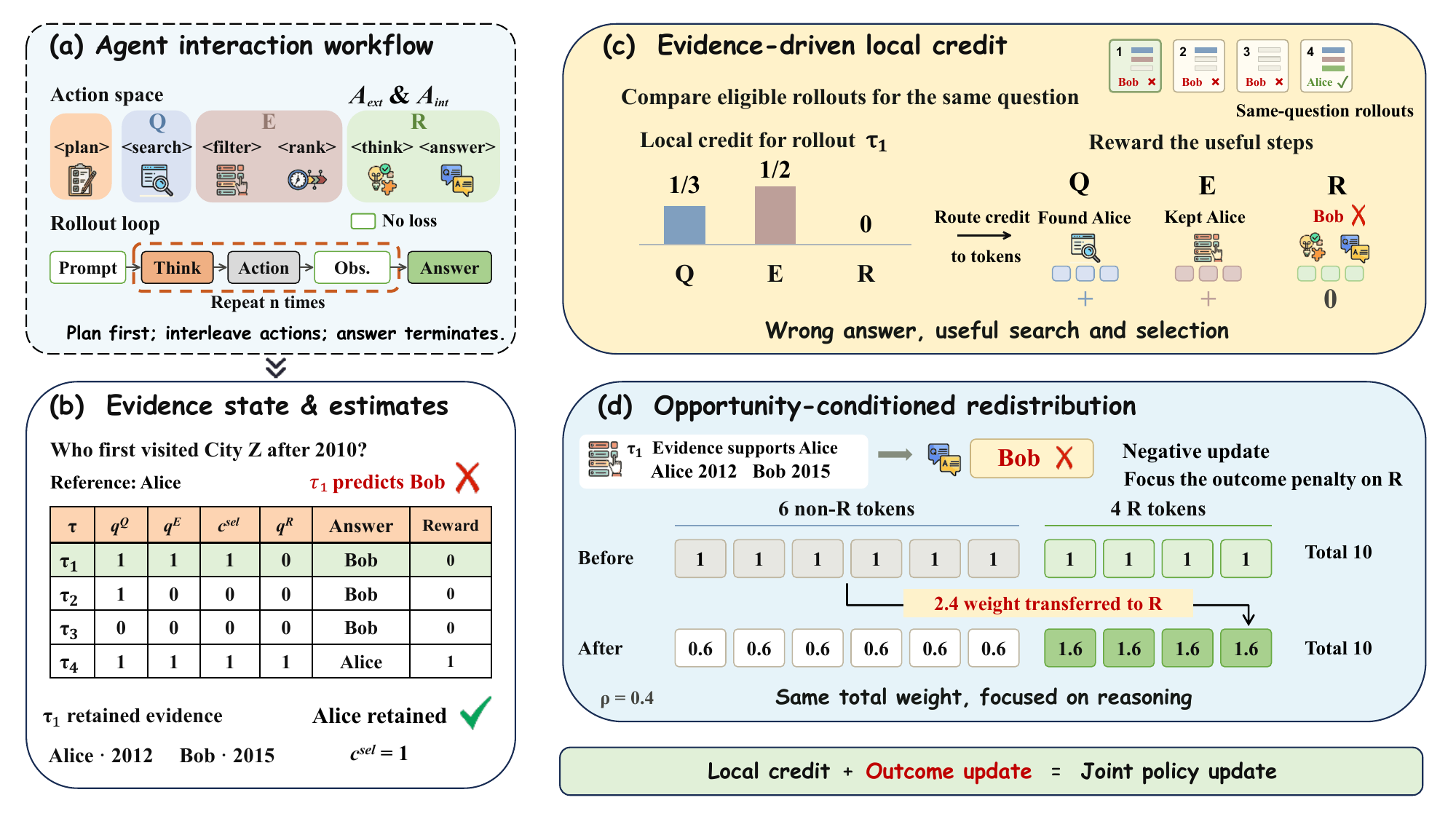}
\caption{Framework of CREDO. CREDO evaluates retrieved and retained evidence to recover local credit through within-question comparisons and redistribute outcome-policy weight toward reasoning and answering. The illustration uses $\rho=0.4$; the experimental default is $\rho=0.04$.}
\label{fig:credo-overview}
\end{figure}

\section{Related Work}
\label{sec:related_work}

\subsection{Temporal Knowledge Graph Question Answering}

Research on factual knowledge spans its acquisition, completion, and updating. KnowPrompt incorporates semantic knowledge from relation labels into prompt tuning for relation extraction~\citep{chen2022knowprompt}, while MKGformer integrates visual and textual representations through multi-level fusion for multimodal knowledge graph completion~\citep{chen2022hybrid}. MultiMedBench examines knowledge updating in medical visual question answering by evaluating the reliability, generality, and locality of knowledge editing~\citep{wen2026multimedbench}. Complementing these studies on acquiring and maintaining knowledge, TKGQA investigates how to use time-dependent facts to answer questions under temporal constraints. Early TKGQA approaches include embedding-based methods that score candidate answers~\citep{saxena2021temporal,mavromatis2022tempoqr,chen2023multitq,chen2022SubGTR}, semantic parsing methods that generate executable queries~\citep{jia2018TEQUILA,neelam2021sygma,chen2024Prog-TQA,ding2022SF-TQA}, and graph-based methods that model structural dependencies~\citep{jia2021EXAQT,sharma2022TwiRGCN,liu2023LGQA,liu2023TMA}. Recent LLM-based methods improve question interpretation through rewriting in TimeR$^{4}$~\citep{qian2024timer4} and decomposition in RTQA~\citep{gong-etal-2025-rtqa}, organize reasoning through temporal tools in TempAgent~\citep{qianyihu-etal-2025-TempAgent} and planning in PoK~\citep{qian2025poK}, and reuse successful solution paths in MemoTime~\citep{tan2026memotime}. Temp-R1 further explores end-to-end reinforcement learning for autonomous temporal reasoning~\citep{gong2026tempr1}. CREDO thus studies how intermediate evidence guides credit assignment during temporal-agent training, considering dependencies among retrieval, selection, and answering.

\subsection{Search Agents and Reinforcement Learning}

PPO~\citep{PPO} and GRPO~\citep{shao2024deepseekmath} optimize policies from outcome rewards without token-level process annotations. OpenAI-o1 and DeepSeek-R1 demonstrate effective reinforcement learning for reasoning~\citep{openai2024openaio1card,guo2025deepseek-R1}, while Search-R1, R1-Searcher, and ReSearch extend it to agents interacting with external knowledge~\citep{jin2025searchr1,song2025r1searcher,chen2025research}. Recent methods supplement terminal feedback with finer-grained credit for intermediate decisions~\citep{feng2025gigpo,ma-etal-2026-tspo}. TIPS derives turn-level rewards from changes in answer likelihood under a teacher model~\citep{xie2026tips}, while StepSearch uses intermediate search rewards and token-level supervision using information gain and redundancy penalties~\citep{wang2025stepsearch}. CREDO instead uses answer-overlap proxies for within-question stage comparisons and evidence-conditioned outcome-weight redistribution, preserving terminal rewards and trajectory-level advantages without an additional learned evaluator. 

\begin{figure}[!t]
\centering
\begin{minipage}{\linewidth}
\small
\refstepcounter{credoalgorithm}\label{alg:credo}
\hrule height 0.8pt
\vspace{4pt}
\noindent\textbf{\credoalgorithmname~\thecredoalgorithm\quad CREDO training update}
\par\vspace{4pt}\hrule height 0.4pt\vspace{4pt}
\noindent\textbf{Input:} Mini-batch $\mathcal B$, behavior policy $\pi_{\mathrm{old}}$, rollout count $K$, and CREDO hyperparameters.
\par\vspace{3pt}
\renewcommand{\arraystretch}{1.06}
\begin{tabular}{@{}r@{\hspace{7pt}}p{\dimexpr\linewidth-20pt\relax}@{}}
1 & \textbf{for each} $(x,G)\in\mathcal B$ \textbf{do} \\
2 & \quad Sample $K$ trajectories with $\pi_{\mathrm{old}}$; compute original rewards and GRPO advantages $\widehat A_i$. \\
3 & \quad Extract evidence and stage masks; determine token availability $t_i^s$. \\
4 & \quad Compute $q_i^Q$, $c_i^{\mathrm{sel}}$, $q_i^E$, and $q_i^R$ using Eqs.~(\ref{eq:evidence-coverage})--(\ref{eq:stage-quality}). \\
5 & \quad Determine comparison validity $v_i^s$; set $\omega_i^E=q_i^Q$ and $\omega_i^R=c_i^{\mathrm{sel}}$. \\
6 & \quad \textbf{for each} stage $s\in\{Q,E,R\}$ \textbf{do} \\
7 & \qquad Compute peer comparisons $F_i^s$ and local advantages $a_i^s$ (Eqs.~(\ref{eq:local-frontier})--(\ref{eq:local-credit})). \\
8 & \qquad Compute the unnormalized stage budget; set it to zero if no positives exist (Appendix~\ref{app:objective}). \\
9 & \quad \textbf{end for} \\
10 & \quad Normalize budgets across stages to obtain $\alpha_s$; allocate $u_{it}^s$ by Eq.~(\ref{eq:local-token-weight}). \\
11 & \quad Identify eligible trajectories (Section~\ref{sec:outcome-redistribution}); assign $w_{it}$ by Eq.~(\ref{eq:redistribution}). \\
12 & \textbf{end for} \\
13 & Optimize $J_{\mathrm{outcome}}^w+\lambda_{\mathrm{local}}J_{\mathrm{local}}$ using separately normalized clipped surrogates; omit zero-weight terms. \\
\end{tabular}
\par\vspace{4pt}\hrule height 0.8pt
\end{minipage}
\end{figure}

\section{Method}
\label{sec:method}

\textbf{Credit Redistribution with Evidence-Driven Opportunities (CREDO)} uses acquired and retained evidence to reward useful intermediate behavior through within-question comparisons and redistribute outcome supervision (Figure~\ref{fig:credo-overview}). Both components jointly optimize one agent while preserving terminal rewards and GRPO trajectory advantages. Algorithm~\ref{alg:credo} summarizes the update.

\subsection{Problem Formulation}
\label{sec:formulation}

Given a temporal knowledge graph $\mathcal K$ and a training pair $(x,G)\in\mathcal D$, an agent $\pi_\theta$ interacts with $\mathcal K$ to answer $x$, where $G$ contains acceptable answers. The behavior policy $\pi_{\mathrm{old}}$ samples $K$ trajectories $\{\tau_i\}_{i=1}^K$ per question. Their original terminal rewards determine the GRPO advantages $\widehat A_i$.

We distinguish retrieval $Q$ (Search), selection $E$ (Filter and Rank), and reasoning/answering $R$ (post-retrieval Think and Answer), which may recur across rounds. Let $A_{it}\in\{0,1\}$ indicate trainable model tokens and $M_{it}^s\leq A_{it}$ their membership in stage $s\in\{Q,E,R\}$. Stage masks are disjoint, exclude environment observations (Information), and locate credit; evidence states determine its assignment. Retrieval and selection masks can aggregate repeated actions; reasoning uses the final post-retrieval Think region and Answer contents. Appendix~\ref{app:processing} specifies localization.

\subsection{Evidence-Driven Local Credit}
\label{sec:local-credit}

\noindent\textbf{Evaluating acquired and retained evidence.} $I_i$ contains all retrieved Information lines. Final selection $S_i$ takes non-empty lines from the last complete Rank output, or the last complete Filter output if Rank is empty or absent. This answer-independent preference reflects Rank's organizational role. Line-level provenance checks yield ordered retained evidence $\widetilde S_i=(l_0,l_1,\ldots)$.

To estimate answer-related evidence, let $\operatorname{cov}(l,g)$ denote the fraction of normalized answer tokens in $g$ covered by fact line $l$, using multiset overlap. Retrieved and retained coverage are
\begin{equation}
q_i^Q=\max_{l\in I_i,\,g\in G}\operatorname{cov}(l,g),
\qquad
c_i^{\mathrm{sel}}=\max_{l_j\in\widetilde S_i,\,g\in G}\frac{\operatorname{cov}(l_j,g)}{\log_2(j+2)}.
\label{eq:evidence-coverage}
\end{equation}
The positional discount gives greater credit to evidence placed earlier in the selection. Empty or unreliable evidence receives zero coverage. We then evaluate selection relative to its retrieved input and answering relative to the references:
\begin{equation}
q_i^E=
\begin{cases}
\min\{c_i^{\mathrm{sel}}/q_i^Q,1\}, & q_i^Q>\delta,\\
0, & \text{otherwise},
\end{cases}
\qquad
q_i^R=\max_{g\in G}\operatorname{F1}_{\mathrm{token}}(\widehat y_i,g),
\label{eq:stage-quality}
\end{equation}
where $\delta>0$ is a numerical threshold and $\widehat y_i$ is the locally extracted answer. Missing or unreliable answers receive zero quality. These coverage and overlap scores serve as local quality proxies; normalization, provenance, and extraction details are provided in Appendix~\ref{app:processing}.

\noindent\textbf{Accounting for evidence dependencies.} We estimate evidence opportunity through upstream answer overlap:
\begin{equation}
\omega_i^E=q_i^Q,
\qquad
\omega_i^R=c_i^{\mathrm{sel}}.
\label{eq:evidence-opportunity}
\end{equation}
Here $\omega_i^E$ measures retrieved overlap and $\omega_i^R$ retained overlap. High retention quality $q_i^E$ can coexist with low $\omega_i^R$ when little answer-related evidence was retrieved. Retrieval has no upstream dependency. These proxies do not verify relations, temporal constraints, or multi-fact completeness: unrelated facts may contain answer tokens, while evidence for computed answers may lack them. Provenance verifies the retrieved source rather than semantic sufficiency; controlled evidence perturbations illustrate this boundary (Appendix~\ref{app:limitations}).

\noindent\textbf{Recovering useful local behavior.} We compare stage quality among trajectories of the same question, allowing intermediate differences to remain informative when terminal rewards coincide. Let $v_i^s\in\{0,1\}$ indicate comparison validity and $\mathcal V_s=\{i:v_i^s=1\}$. For $i\in\mathcal V_s$ and $|\mathcal V_s|>1$, define
\begin{equation}
F_i^s=\frac{1}{|\mathcal V_s|-1}\sum_{j\in\mathcal V_s\setminus\{i\}}\operatorname{sgn}(q_i^s-q_j^s).
\label{eq:local-frontier}
\end{equation}
Ties contribute zero, so a positive score indicates more wins than losses against valid peers. We set $F_i^s=0$ when comparison is unavailable. With $[z]_+=\max(z,0)$ and $t_i^s\in\{0,1\}$ indicating reliable stage-token localization, the auxiliary advantages are
\begin{equation}
a_i^Q=v_i^Qt_i^Qq_i^Q[F_i^Q]_+,
\qquad
a_i^s=v_i^st_i^s\omega_i^s q_i^s[F_i^s]_+,
\quad s\in\{E,R\}.
\label{eq:local-credit}
\end{equation}
\begin{wrapfigure}{r}{0.42\textwidth}
\vspace{-7pt}
\centering
\includegraphics[width=\linewidth]{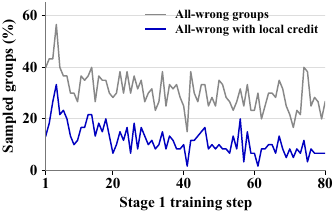}
\caption{\small All-wrong groups and those with active local credit in CREDO Stage~1 (8B). Both denominators count all sampled groups.}
\label{fig:all-failed-coverage}
\vspace{-7pt}
\end{wrapfigure}
Locally superior behavior receives positive credit, scaled by upstream opportunity for dependent stages. Evaluable failures can be zero-quality peers without trainable tokens. Local quality differences can thus supply positive supervision even in all-failed groups.

\noindent\textbf{Empirical illustration.} Figure~\ref{fig:all-failed-coverage} shows recurring all-wrong groups during Stage~1. Of these, 38.29\% retain active local credit despite zero terminal GRPO advantages. Activity requires positive local advantages, trainable tokens, and allocated weights. The trace demonstrates signal availability, without isolating its contribution to final accuracy. Counts and reward settings appear in Appendix~\ref{app:all-wrong}.

\CREDOclearwrap

\begin{table}[t]
\caption{Performance comparison on MultiTQ. CREDO follows the normalized substring-matching protocol in Temp-R1's public evaluation code. Baseline scores are adopted from Table~3 of Temp-R1~\citep{gong2026tempr1}. The \protect\bestscore{best} and \protect\secondscore{second-best} scores under each metric are highlighted in colors, respectively; ties receive the same highlight.}
\label{tab:multitq-main}
\centering
\small
\setlength{\tabcolsep}{4pt}
\begin{tabular*}{\linewidth}{@{\extracolsep{\fill}}lccccc@{}}
\toprule
\multirow{2}{*}{Model} & \multirow{2}{*}{Overall} & \multicolumn{2}{c}{Question Type} & \multicolumn{2}{c}{Answer Type} \\
\cmidrule(lr){3-4}\cmidrule(lr){5-6}
& & Multiple & Single & Entity & Time \\
\midrule
\multicolumn{6}{c}{\textit{TKG Embedding-based methods}} \\
\midrule
EmbedKGQA & 0.206 & 0.134 & 0.235 & 0.290 & 0.001 \\
CronKGQA & 0.279 & 0.134 & 0.337 & 0.328 & 0.156 \\
MultiQA & 0.293 & 0.159 & 0.347 & 0.349 & 0.157 \\
\midrule
\multicolumn{6}{c}{\textit{Prompt-based LLMs}} \\
\midrule
ARI & 0.380 & 0.210 & 0.680 & 0.394 & 0.344 \\
TempAgent & 0.702 & 0.316 & 0.857 & 0.624 & 0.870 \\
MemoTime & 0.730 & 0.459 & 0.829 & 0.677 & 0.846 \\
RTQA & 0.765 & 0.424 & 0.902 & 0.692 & 0.942 \\
\midrule
\multicolumn{6}{c}{\textit{Fine-tuning-based LLMs}} \\
\midrule
Search-R1 & 0.352 & 0.094 & 0.474 & 0.230 & 0.705 \\
TimeR4 & 0.728 & 0.335 & 0.887 & 0.639 & 0.945 \\
PoK & 0.779 & 0.409 & \bestscore{0.929} & 0.696 & 0.962 \\
\textbf{Temp-R1} & \secondscore{0.780} & \secondscore{0.550} & 0.888 & \secondscore{0.714} & \secondscore{0.969} \\
\textbf{CREDO} & \bestscore{0.824} & \bestscore{0.656} & \secondscore{0.903} & \bestscore{0.771} & \bestscore{0.977} \\
\bottomrule
\end{tabular*}
\end{table}

\subsection{Evidence-Conditioned Outcome Redistribution}
\label{sec:outcome-redistribution}

\noindent\textbf{Locating outcome supervision.} The original negative GRPO advantage can still penalize useful evidence acquisition. When retained answer overlap indicates an answering opportunity, we therefore shift part of the outcome-policy weight toward reasoning and answering.

Let $n_i=\sum_t A_{it}$ and $m_i=\sum_t M_{it}^R$ be the numbers of trainable and reasoning/answering tokens. Redistribution requires $v_i^R=1$, $\omega_i^R>0$, $n_i,m_i>0$, and $m_i/n_i\geq\eta$, where $\eta\in(0,1]$ prevents excessive concentration on a short token span. For eligible trajectories $\mathcal E$, we assign
\begin{equation}
w_{it}=
\begin{cases}
(1-\rho)A_{it}+\rho\dfrac{n_i}{m_i}M_{it}^R, & i\in\mathcal E,\\[4pt]
A_{it}, & i\notin\mathcal E,
\end{cases}
\qquad
\sum_t w_{it}=n_i.
\label{eq:redistribution}
\end{equation}
Here $\rho\in[0,1)$ mixes the original token weights with a reasoning/answering-only allocation; the net weight moved into this region is $\rho(n_i-m_i)$. The factor $n_i/m_i$ preserves total policy weight. Evidence opportunity determines eligibility rather than continuously scaling $\rho$. The conserved quantity is the sum of policy weights, not the gradient magnitude.

Redistribution applies to either sign of $\widehat A_i$. When $0<\rho$ and $m_i<n_i$, it increases reasoning/answering weights and decreases others; if $m_i=n_i$, weights are unchanged. If $\widehat A_i=0$, redistribution creates no outcome signal, but local credit can exploit intermediate quality differences.

\subsection{Policy Optimization}
\label{sec:optimization}

\noindent\textbf{Normalizing local supervision.} To keep auxiliary learning from scaling with trajectory length or the number of positive examples, we assign each stage a normalized budget $\alpha_s$ within each question group (the group index is omitted here). This budget combines valid comparison support, upstream opportunity for dependent stages, within-group quality diversity, and remaining improvement; its complete definition appears in Appendix~\ref{app:objective}. Let $\mathcal H_s=\{i:a_i^s>0\}$, $h_s=|\mathcal H_s|$, and $n_i^s=\sum_t M_{it}^s$. For each positive example, define
\begin{equation}
u_{it}^s=\frac{\alpha_s}{h_s n_i^s}M_{it}^s,
\qquad i\in\mathcal H_s,
\qquad
\sum_{i,t}u_{it}^s=\alpha_s.
\label{eq:local-token-weight}
\end{equation}
Other token weights are zero; stages without valid positive examples receive zero budget. The allocation controls total stage weight, while advantage magnitudes still determine the weighted loss.

\noindent\textbf{Joint optimization.} We combine the redistributed outcome signal with the recovered local credit:
\begin{equation}
J(\theta)=J_{\mathrm{outcome}}^w(\theta)+\lambda_{\mathrm{local}}J_{\mathrm{local}}(\theta),
\label{eq:credo-objective}
\end{equation}
where $J_{\mathrm{outcome}}^w$ uses the original GRPO advantage $\widehat A_i$ with token weights $w_{it}$, and $J_{\mathrm{local}}$ uses stage advantages $a_i^s$ with weights $u_{it}^s$. The outcome term uses token-mean aggregation within each static microbatch, with redistributed token weights in both the numerator and denominator. The local term uses PPO-minibatch weight normalization; Appendix~\ref{app:objective} specifies clipping, accumulation, and regularization. Zero-weight terms are omitted. The coefficient $\lambda_{\mathrm{local}}$ controls auxiliary supervision. Local credit determines which stage behavior to reinforce; redistribution determines where outcome supervision acts. Inference requires neither reference answers nor credit computations.

\begin{table}[!t]
\caption{Performance comparison on TimelineKGQA across in-domain (CronQuestion KG) and out-of-domain (ICEWS Actor) settings. $^{*}$Our reproduction, using the same base training configuration as CREDO and Temp-R1's public scoring protocol (Appendix~\ref{app:training}). Other baseline results are adopted from Temp-R1~\citep{gong2026tempr1}. The \protect\bestscore{best} and \protect\secondscore{second-best} scores under each metric are highlighted in colors, respectively; ties receive the same highlight.}
\label{tab:timeline-main}
\centering
\small
\resizebox{\linewidth}{!}{%
\begin{tabular}{lcccccccc}
\toprule
\multirow{2}{*}{Model} & \multicolumn{4}{c}{CronQuestion KG (In Domain)} & \multicolumn{4}{c}{ICEWS Actor (Out of Domain)} \\
\cmidrule(lr){2-5}\cmidrule(lr){6-9}
& Overall & Simple & Medium & Complex & Overall & Simple & Medium & Complex \\
\midrule
RAG Baseline & 0.235 & \secondscore{0.704} & 0.092 & 0.009 & 0.265 & \bestscore{0.660} & \secondscore{0.128} & 0.011 \\
LLaMA2-7B & 0.169 & 0.049 & 0.143 & 0.282 & 0.111 & 0.035 & 0.066 & 0.322 \\
GPT-4o & 0.206 & 0.069 & 0.130 & 0.376 & 0.113 & 0.051 & 0.035 & \secondscore{0.353} \\
RTQA & 0.298 & 0.608 & 0.218 & 0.135 & -- & -- & -- & -- \\
Temp-R1$^{*}$ & \secondscore{0.508} & \bestscore{0.844} & \secondscore{0.241} & \secondscore{0.446} & \secondscore{0.309} & 0.447 & 0.123 & 0.324 \\
\textbf{CREDO} & \bestscore{0.523} & \bestscore{0.844} & \bestscore{0.278} & \bestscore{0.456} & \bestscore{0.364} & \secondscore{0.501} & \bestscore{0.170} & \bestscore{0.393} \\
\bottomrule
\end{tabular}%
}
\end{table}

\section{Experiments}
\label{sec:experiments}

\subsection{Setup}
\label{sec:experimental-setup}

\noindent\textbf{Datasets.} We train CREDO on two datasets: MultiTQ~\citep{chen2023multitq} and TimelineKGQA-CRON~\citep{timeline}. We evaluate three settings: MultiTQ, CronQuestion KG, and ICEWS Actor. Following Temp-R1's evaluation layout~\citep{gong2026tempr1}, we report MultiTQ results by question and answer type, and evaluate both in-domain question answering and cross-KG transfer on TimelineKGQA. Tables~\ref{tab:multitq-main} and~\ref{tab:timeline-main} present these comparisons. Section~\ref{sec:benchmarks-metrics} describes the tasks and evaluation protocol; dataset sizes, category distributions, and data-split audits appear in Appendix~\ref{app:training}.

\noindent\textbf{Baselines.} On MultiTQ, we compare CREDO with three baseline groups: (1) TKG embedding-based methods, including EmbedKGQA~\citep{embedkgqa}, CronKGQA~\citep{saxena2021temporal}, and MultiQA~\citep{chen2023multitq}; (2) prompt-based LLM methods, including ARI~\citep{ari}, TempAgent~\citep{qianyihu-etal-2025-TempAgent}, MemoTime~\citep{tan2026memotime}, and RTQA~\citep{gong-etal-2025-rtqa}; and (3) fine-tuning-based LLM methods, including Search-R1~\citep{jin2025searchr1}, TimeR4~\citep{qian2024timer4}, PoK~\citep{qian2025poK}, and Temp-R1~\citep{gong2026tempr1}. TimelineKGQA baselines include RAG Baseline, LLaMA2-7B, GPT-4o, RTQA, and Temp-R1. For TimelineKGQA, we train Temp-R1 and CREDO under identical base configurations and evaluate them on identical inputs using Temp-R1's public scoring protocol (Appendix~\ref{app:training}), reducing confounding from protocol differences and making the comparison more directly reflect differences between methods. Other baseline scores come from Tables~3 and~2 of Temp-R1 for MultiTQ and TimelineKGQA, respectively.

\noindent\textbf{Implementation Details.} We use Meta-Llama-3.1-8B-Instruct with E5-base-v2 retrieval and a FAISS Flat index, following \citet{jin2025searchr1}. Training follows Temp-R1's two-stage procedure~\citep{gong2026tempr1}. CREDO-family runs share initial assets and base training settings; selected models use a common evaluation configuration. Checkpoints are selected by monitoring accuracy (Appendix~\ref{app:training}). Defaults are $\lambda_{\mathrm{local}}=0.03$, $\rho=0.04$, and $\eta=0.05$. All 8B reinforcement learning experiments run on one machine with 8 Ascend 910B NPUs. Appendix~\ref{app:training} gives the configurations and distinguishes these ablations from subsequent frozen-model diagnostics.

\begin{figure}[!t]
\centering
\includegraphics[width=\linewidth]{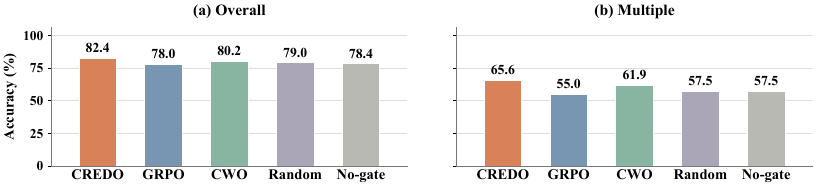}
\caption{MultiTQ component comparisons: (a) Overall; (b) Multiple. GRPO uses the Temp-R1 reported results~\citep{gong2026tempr1}. CWO disables redistribution; Random changes its target tokens; No-gate adds ungated answer-region redistribution to CWO. See Appendix~\ref{app:training} for selection details.}
\label{fig:ablation}
\end{figure}

\subsection{Benchmarks and Metrics}
\label{sec:benchmarks-metrics}

\noindent\textbf{Temporal KGQA.} Given a question and a temporal knowledge graph, the agent returns the requested entity, time, or other value under the question's temporal constraints. Evaluation concerns final answers without requiring a predefined tool-use trajectory, allowing different retrieval and reasoning paths across one or more interaction rounds to produce acceptable answers.

\noindent\textbf{MultiTQ.} MultiTQ~\citep{chen2023multitq} evaluates temporal reasoning at year, month, and day granularities. The official Single categories are Equal, Before/After, and First/Last; Multiple includes Equal Multi, Before Last, and After First. We use these official labels without treating them as measured constraint counts. Our fixed evaluation set contains 500 questions, comprising 340 Single and 160 Multiple questions. Table~\ref{tab:multitq-main} reports results by question type and answer type.

\noindent\textbf{TimelineKGQA.} TimelineKGQA~\citep{timeline} categorizes questions as Simple, Medium, or Complex according to whether they require one, two, or three contextual facts. Tasks include temporally constrained retrieval, comparison, composition, and ranking; answers include entities, time points, intervals, durations, and temporal relations. Following Temp-R1, we evaluate on CronQuestion KG and directly transfer the same checkpoint to ICEWS Actor without further training, assessing cross-KG transfer. Table~\ref{tab:timeline-main} reports Overall and complexity-level results.

\noindent\textbf{Metrics.} Following the evaluation protocol in the public Temp-R1 repository~\citep{gong2026tempr1}, we evaluate final answers using normalized substring matching. A prediction is correct if any acceptable answer occurs contiguously within the normalized final answer. Let $\hat{y}_i$ denote the predicted answer, $G_i$ the set of acceptable answers, and $\nu(\cdot)$ the normalization function. Accuracy over $N$ questions is defined as:
\begin{equation}
\operatorname{Acc}_{\mathrm{sub}}=\frac{1}{N}\sum_{i=1}^{N}\mathbf{1}\!\left[\exists\,g\in G_i:\operatorname{substr}\!\left(\nu(g),\nu(\hat{y}_i)\right)\right],
\label{eq:substring}
\end{equation}
where $\operatorname{substr}(a,b)$ indicates contiguous occurrence. Answers are extracted from \texttt{<answer>} tags; normalization lowercases text, removes punctuation and English articles, and collapses whitespace. Missing answers and execution failures count as incorrect. This metric follows the baseline protocol and measures answer containment; it does not require the output to contain only a single unambiguous answer. Appendix~\ref{app:processing} provides extraction details.

\subsection{Experimental Results}
\label{sec:results}

\noindent\textbf{Results on MultiTQ.} CREDO achieves the best Overall and Multiple accuracy in Table~\ref{tab:multitq-main}, exceeding the strongest published baseline by 4.4 and 10.6 percentage points, respectively. The larger gain on Multiple questions highlights its effectiveness on compositional temporal tasks, where agents must coordinate evidence acquisition, selection, and answer derivation. This pattern aligns with the design of CREDO: local comparisons recover supervision from intermediate progress, while evidence-conditioned redistribution directs outcome-policy weight toward reasoning and answering. These mechanisms address the dependencies that make useful retrieval alone insufficient for solving complex questions. CREDO also ranks first on both entity and time answers, showing that its performance advantage extends across answer formats.

\noindent\textbf{Results on TimelineKGQA.} In Table~\ref{tab:timeline-main}, CREDO improves Overall accuracy over reproduced Temp-R1 on CronQuestion KG and direct ICEWS Actor transfer by 1.5 and 5.5 percentage points, respectively. On CronQuestion KG, the gains come from Medium and Complex questions while Simple accuracy is maintained, highlighting the benefit on tasks requiring multiple contextual facts to be integrated under the question's temporal constraints. The larger improvement on ICEWS Actor, with gains across all three complexity levels, further supports transfer beyond the training knowledge graph. The same Cron-trained checkpoint is applied directly without additional training, demonstrating that the learned policy remains effective in a new graph environment. Under matched training and evaluation settings, these results support evidence-conditioned credit assignment for both compositional temporal question answering and direct cross-KG transfer.

\begin{figure}[!t]
\centering
\includegraphics[width=\linewidth]{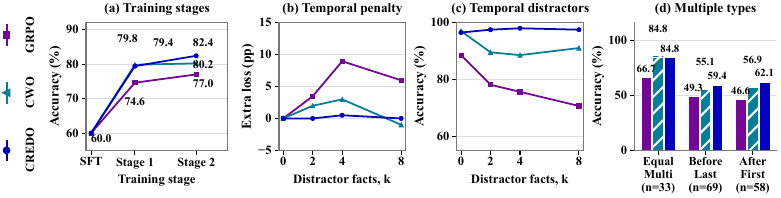}
\caption{CREDO diagnostics. (a) Stage-wise accuracy on 500 MultiTQ inputs; the GRPO Stage~2 score is from our reproduction under the same training and evaluation settings as CREDO. (b) Additional substring-score loss from temporal rather than generic distraction. (c) Substring accuracy with temporal distractors, averaged over three decoding seeds on 67 questions. (d) Official Multiple types from the paired evaluation, with group sizes below the axis. Sources: Appendix~\ref{app:versions}.}
\label{fig:credo-diagnostics}
\label{fig:dynamics-cost}
\label{fig:temporal-robustness}
\end{figure}

\subsection{Analysis}
\label{sec:analysis}

\noindent\textbf{Performance across training stages.} Figure~\ref{fig:credo-diagnostics}(a) summarizes the stage-wise evaluation records. CREDO achieves the highest final-stage accuracy, followed by CWO and our reproduced GRPO baseline. The matched final-stage comparison is consistent with a benefit from local credit and an additional improvement from outcome redistribution. Earlier intermediate-stage records are retained as reported; evaluation details appear in Appendix~\ref{app:versions}.

\noindent\textbf{Contribution of local credit and outcome redistribution.} CWO (CREDO without outcome redistribution) retains local credit, allowing Figure~\ref{fig:ablation} to examine the additional value of redistribution under the same two-stage schedule, selection criterion, and evaluation settings. CREDO improves Overall and Multiple accuracy by 2.2 and 3.75 percentage points, respectively, corresponding to net gains of 11 and six correct answers. Because both methods already recover local supervision, these observed differences suggest that controlling where outcome supervision acts can provide further value beyond local credit alone. The larger gain on Multiple questions is consistent with the motivation for emphasizing reasoning and answering in evidence-dependent tasks. Appendix~\ref{app:paired} reports the paired estimates and their uncertainty to contextualize the observed performance differences.

\noindent\textbf{Redistribution variants.} Random preserves eligibility, transferred mass, and target-token count; No-gate retains local credit and increases answer-region weights for trajectories satisfying structural and length conditions, without checking evidence overlap. CREDO exceeds both on Overall and Multiple under the common selection and evaluation protocol (Figure~\ref{fig:ablation}), favoring its evidence-conditioned configuration for determining where outcome supervision should act.

\noindent\textbf{Official Multiple question types.} In Figure~\ref{fig:credo-diagnostics}(d), CREDO exceeds CWO on Before Last and After First, matches it on Equal Multi, and exceeds in-house GRPO throughout. These labels denote question types, not constraint counts (Appendix~\ref{app:paired}).

\noindent\textbf{Comparison under matched settings.} Under identical base training and evaluation settings, CREDO exceeds our Temp-R1 reproduction by 5.4 percentage points in Overall accuracy and 13.75 points on Multiple questions. The improvement also holds against a baseline evaluated within our setup. The larger Multiple gain supports evidence-conditioned training's effectiveness on compositional temporal tasks. Appendix~\ref{app:paired} reports paired results and confidence intervals.

\noindent\textbf{Answer containment under temporal distraction.} Retaining supporting facts without retrieval, CREDO shows near-constant substring accuracy on 67 Single entity questions (Fig.~\ref{fig:credo-diagnostics}(c)). On 60 Multiple questions, it leads in three evidence conditions despite CWO's smaller temporal-distractor drop. Both use three decoding seeds (construction and results: Appendix~\ref{app:evidence-tests}).

\noindent\textbf{Temporal-specific penalty.} On the Single subset, Figure~\ref{fig:credo-diagnostics}(b) compares generic and temporal distractors at matched counts and lengths, retaining supporting facts. CREDO combines a near-zero penalty with high substring accuracy (panel~(c)), indicating stable target-answer containment with little additional loss from temporal distraction. GRPO's larger penalty highlights CREDO's advantage under fixed evidence. Negative penalties indicate higher accuracy with temporal distractors.

\begin{wraptable}{r}{0.43\textwidth}
\centering
\caption{Peak Multiple accuracy on MultiTQ (3B), using the same evaluation settings as the 8B experiments. The default configuration is shown in bold.}
\label{tab:sensitivity}
\small
\setlength{\tabcolsep}{7pt}
\begin{tabular*}{\linewidth}{@{\extracolsep{\fill}}rrr@{}}
\toprule
$\lambda_{\mathrm{local}}$ & $\rho$ & Multiple (\%) \\
\midrule
\textbf{0.03} & \textbf{0.04} & \textbf{45.00} \\
\addlinespace[2pt]
0.01 & 0.04 & 43.75 \\
0.05 & 0.04 & 45.00 \\
\addlinespace[2pt]
0.03 & 0.02 & 45.00 \\
0.03 & 0.06 & 45.62 \\
\bottomrule
\end{tabular*}
\end{wraptable}

\noindent\textbf{Hyperparameter sensitivity.} Table~\ref{tab:sensitivity} examines $\lambda_{\mathrm{local}}$ and $\rho$ using Llama-3.2-3B-Instruct, varying one coefficient at a time while keeping all other training and evaluation settings fixed. For every setting, we evaluate the saved candidates on the same official 500-question MultiTQ set and report the highest accuracy on its 160 Multiple questions. The reported peaks are close across the tested values, indicating comparable attainable performance around the default configuration within this training budget. This exploratory comparison concerns peak performance rather than performance under the monitoring-based selection procedure used for the main 8B experiments.

\CREDOclearwrap

\section{Conclusion}
\label{sec:conclusion}

In this paper, we introduce CREDO for evidence-conditioned credit assignment in temporal reasoning agents. CREDO recovers positive local supervision through same-question comparisons and reallocates outcome-policy weight toward reasoning and answering based on retained answer overlap. It preserves terminal rewards, GRPO advantages, and total outcome-policy weight without an additional learned evaluator or inference-time credit computation. Results on MultiTQ and TimelineKGQA show gains in compositional question answering and cross-KG transfer, while component comparisons support the combined training procedure under a common evaluation protocol.

\bibliography{iclr2027_conference}
\bibliographystyle{iclr2027_conference}

\clearpage
\appendix
\section{Additional Details and Results}
\label{app:details}

\subsection{Implementation Details}
\label{app:processing}
\label{app:objective}

\noindent\textbf{Evidence and answer processing.} Local scoring lowercases text, replaces ASCII punctuation with spaces, removes English articles, and collapses whitespace. Coverage is the fraction of reference-answer tokens found in a fact line, using multiset overlap; local answer F1 uses the same normalization. Retrieved evidence pools all complete Information blocks. The retained selection uses the last complete Rank block, falling back to the last complete Filter when that Rank is absent or empty. Document prefixes and title wrappers are removed before checking normalized selected lines against the retrieved-line set. Untraceable lines are discarded individually, and surviving lines are reindexed before positional discounting. Duplicate selected lines are retained. This response-wide provenance check verifies the retrieved source, not semantic or temporal sufficiency.

The local answer extractor accepts complete Answer tags case-insensitively and selects the last match shorter than 200 characters, or the shortest match if all are longer. The archived terminal extractor accepts lowercase tags and, when any short match exists, returns the last match regardless of length; otherwise it returns the shortest. Terminal normalization deletes punctuation rather than replacing it with spaces. The two branches can therefore evaluate different extracted or normalized text when outputs contain multiple Answer blocks. Missing answers receive zero quality or score.

\noindent\textbf{Validity and token localization.} Let $T_i$ indicate truncation, $C_i^Q$ reliable Search/Information structure with an Information response before the next Search, $C_i^E$ reliable Filter/Rank structure, and $N_i^E$ non-empty selection content. The stage-local rules are
\begin{align}
 v_i^Q&=C_i^Q\lor\neg T_i,\notag\\
 v_i^E&=v_i^Q\land(q_i^Q>\epsilon_0)\land[\neg T_i\lor(C_i^E\land N_i^E)],\notag\\
 v_i^R&=v_i^E\land(c_i^{\mathrm{sel}}>\epsilon_0)\land\neg T_i,
 \qquad\epsilon_0=10^{-8}.
 \label{eq:app-validity}
\end{align}
Here $\epsilon_0$ is the coverage threshold $\delta$ in the main text. These validity rules govern local comparisons and the evidence-conditioned CREDO route. No-gate retains them for local credit but uses structural and token-length eligibility for redistribution, without the overlap requirements in $v_i^R$ (Appendix~\ref{app:training}). Separately, $t_i^s$ requires a non-empty reliable token mask. Thus, an evaluable failure may contribute a zero-quality comparison without supplying teacher tokens.

The retrieval mask contains complete Search contents, and the selection mask aggregates complete Filter and Rank contents. The reasoning/answering mask contains complete Think contents starting after the last Information block (or the last Search when Information is absent), together with complete Answer contents, subject to structural validity. Tokens must lie entirely inside their content spans. Tokenizer offsets, with prefix-decoding checks when needed, align text spans to tokens. Global alignment failures, overlapping stage masks, or out-of-bounds masks disable local credit. Plan, tags, and unclassified model text receive no local mask but may retain outcome supervision. Information and padding are excluded from actor updates. These masks aggregate actions across rounds; the credit is stage-level rather than a separate estimate for each action.

\noindent\textbf{Stage budgets.} Budgets are computed per complete question group $g$ before PPO minibatch partitioning. Write $V_{gs}=|\mathcal V_{gs}|$ for the valid comparison count and $K_g$ for the sampled group size. Let $(o_i^Q,o_i^E,o_i^R)=(1,q_i^Q,c_i^{\mathrm{sel}})$, and let $\mathcal H_{gs}$ contain positive local teachers with non-empty masks. For $V_{gs}\geq2$, define
\begin{align}
 b_{gs}&=V_{gs}/K_g,&
 \bar o_{gs}&=\frac{1}{V_{gs}}\sum_{i\in\mathcal V_{gs}}o_i^s,\notag\\
 d_{gs}&=\frac{\sum_{i,j\in\mathcal V_{gs}}|q_i^s-q_j^s|}{V_{gs}(V_{gs}-1)},&
 e_{gs}&=\left[1-\frac{1}{V_{gs}}\sum_{i\in\mathcal V_{gs}}q_i^s\right]_+.
\end{align}
These factors represent comparison support, opportunity, quality dispersion, and remaining improvement. The budget is
\begin{equation}
 \ell_{gs}=b_{gs}\bar o_{gs}d_{gs}e_{gs}\mathbf1[|\mathcal H_{gs}|>0],
 \qquad
 \alpha_{gs}=\begin{cases}
 \ell_{gs}/\sum_{s'}\ell_{gs'},&\sum_{s'}\ell_{gs'}>\epsilon_0,\\
 0,&\text{otherwise}.
 \end{cases}
 \label{eq:app-budget}
\end{equation}
Incomplete groups, unavailable comparisons, non-finite values, and stages without positive teachers receive zero corresponding credit. Frontier ties contribute zero. Equation~\ref{eq:local-token-weight} then divides each stage budget equally among its teachers and their stage tokens. Budgets are normalized across stages within a question, without additional question-difficulty weighting.

\noindent\textbf{Policy loss and regularization.} GRPO standardizes each trajectory reward within its question group using the sample standard deviation and denominator offset $10^{-6}$. Local credit does not enter this reward. Outcome redistribution uses Equation~\ref{eq:redistribution}, with a $10^{-6}$ tolerance in the minimum token-fraction check and implementation range $0\leq\rho<1$. The net weight transferred from outside the reasoning/answering region is $\rho(n_i-m_i)$.

In minimization form, the PPO implementation uses the clipped ratio $r=\exp(\operatorname{clip}(\log\pi_\theta-\log\pi_{\mathrm{old}},-20,20))$ and dual clipping:
\begin{align}
 \phi(a,r)&=\max\{-ar,-a\operatorname{clip}(r,0.8,1.2)\},\notag\\
 \widetilde\phi(a,r)&=\begin{cases}\min\{\phi(a,r),-3a\},&a<0,\\\phi(a,r),&a\geq0.\end{cases}
\end{align}
The local advantages are non-negative and do not activate the negative-advantage branch. Saved old-policy log probabilities are used for both terms.

Disjoint stage masks define token-level local advantages $a_{it}^{\mathrm{loc}}=\sum_s M_{it}^s a_i^s$ and weights $u_{it}=\sum_su_{it}^s$. For microbatch $B_\mu$, the policy losses are
\begin{align}
L_{\mathrm{out},\mu}
&=
\frac{\sum_{(i,t)\in B_\mu}w_{it}\widetilde{\phi}(\widehat A_i,r_{it})}
{\sum_{(i,t)\in B_\mu}w_{it}},
\notag\\
L_{\mathrm{loc},\mu}
&=
\frac{\sum_{(i,t)\in B_\mu}u_{it}\widetilde{\phi}(a_{it}^{\mathrm{loc}},r_{it})}
{\max(U_\mu,10^{-6})},
\qquad
U_\mu=\sum_{(i,t)\in B_\mu}u_{it}.
\end{align}
Both training stages use the legacy actor implementation with static microbatches and token-mean loss aggregation (\texttt{use\_legacy\_worker\_impl=enable}, \texttt{use\_dynamic\_bsz=False}, and \texttt{loss\_agg\_mode=token-mean}). The executed path receives no global token-count metadata, so the outcome denominator is computed within the current microbatch. Since redistribution preserves each trajectory's total weight, this denominator equals the number of valid actor tokens in that microbatch, excluding tool observations and padding. Redistribution therefore changes relative token weighting without changing the denominator for a fixed microbatch.

For a PPO minibatch, local weight is summed across the default distributed process group to obtain $U_*$, with world size $W$. After accumulation scaling, the loss contributions are
\begin{equation}
 \sum_\mu\left[
 \gamma_\mu\big(L_{\mathrm{out},\mu}+\beta L_{\mathrm{KL},\mu}-\zeta H_\mu\big)
 +\lambda_{\mathrm{local}}\frac{WU_\mu}{U_*}L_{\mathrm{loc},\mu}
 \right].
 \label{eq:app-total-loss}
\end{equation}
The local term is omitted if $U_*=0$. Static microbatches use the reciprocal accumulation count for $\gamma_\mu$; the local branch compensates for that scaling before backward. Stage budgets therefore retain their relative weights inside the local loss. Each PPO minibatch ends with an optimizer update. Outcome losses are averaged through static microbatch accumulation, rather than normalized by a single token count over the entire PPO minibatch.

KL is an independent actor loss and is not added to the checked training rewards. The low-variance estimator is $\operatorname{clip}(\exp(z)-z-1,-10,10)$ for $z=\operatorname{clip}(\log\pi_{\mathrm{ref}}-\log\pi_\theta,-20,20)$. KL and entropy use the original actor response mask, not redistributed or local weights. Weight conservation thus does not imply invariant gradient magnitude.

\subsection{Experimental Setup}
\label{app:training}
\label{app:versions}

\noindent\textbf{Data and training.} The fixed MultiTQ evaluation contains 340 Single and 160 Multiple questions, with 371 entity and 129 time answers. The year/month/day counts are 69/105/326. These are official categories, not inferred constraint counts. The processed data records contain 57,979 MultiTQ training-pool entries and 54,584 test entries, and 24,427 Cron training entries and 8,146 test entries; these describe the local processed files rather than the original benchmark release sizes. Normalized-question overlap between the processed MultiTQ training and test files is zero. The monitoring sets are seed-42 samples: 24 questions from \texttt{test\_multiple} for Stage~1 and 30 questions from \texttt{test\_all} for Stage~2. Both monitoring sets are disjoint from the fixed 500-question evaluation set. Checkpoints for the main comparisons are selected using monitoring accuracy; the 500-question evaluation scores are not used for this selection.

The policy is Meta-Llama-3.1-8B-Instruct, with E5-base-v2 retrieval and a FAISS Flat index. Table~\ref{tab:app-training} gives the common MultiTQ settings for the in-house GRPO, CREDO, CWO, Random, and No-gate runs. Each starts from the same initial SFT weights and continues its own Stage~1 step-80 checkpoint into Stage~2 with a reinitialized optimizer. The main CREDO result uses Stage~2 step~400. Defaults are $\lambda_{\mathrm{local}}=0.03$, $\rho=0.04$, and $\eta=0.05$. CWO is the configuration named CF in archived files: local credit remains active, while outcome redistribution is disabled.

\begin{table}[!htbp]
\centering\small
\caption{Common base training configuration for the five in-house 8B methods on MultiTQ. Generation limits and response-container limits measure different quantities.}
\label{tab:app-training}
\begin{tabular*}{\linewidth}{@{\extracolsep{\fill}}lcc@{}}
\toprule
Setting & Stage 1 & Stage 2\\
\midrule
Scheduled updates & 80 & 500\\
Questions per update $\times$ rollouts & $60\times5$ & $2\times8$\\
PPO minibatch trajectories & 30 & 2\\
Data / sequence parallelism & 1 / 8 & 2 / 4\\
Learning rate & $5\times10^{-7}$ & $5\times10^{-7}$\\
Gradient clipping & 5 & 1\\
KL / entropy coefficients & 0.01 / 0.001 & 0.001 / 0.001\\
Sampling temperature / top-$p$ & 1 / 0.95 & 1 / 0.95\\
Retrieval top-$k$ & 30 & 20\\
Maximum searches / model calls & 3 / 4 & 3 / 4\\
Per-call generation limit & 1,000 & 1,024\\
Configured generation / response limit & 4,000 / 5,200 & 4,096 / 8,192\\
Per-call Information limit & 1,400 & 1,500\\
Monitoring interval / questions & 10 / 24 & 20 / 30\\
\bottomrule
\end{tabular*}
\end{table}

\noindent\textbf{Controlled ablations and model selection.} Both training stages use normalized substring rewards in GRPO, CREDO, CWO, Random, and No-gate. The archived configurations agree on batch sizes, rollout counts, optimizer settings, KL and entropy coefficients, token budgets, retrieval settings, and scheduled updates. The CREDO-family Stage~1 input manifests record identical hashes for 12 assets, including all four SFT weight shards, processed data files, the knowledge graph, and the retrieval index. GRPO uses outcome-only supervision; CWO removes outcome redistribution; Random changes its target positions while preserving eligibility, target count, and transferred mass. No-gate retains CWO's local-credit update. Its redistribution route requires reliable reasoning/answering-token localization, $n_i,m_i>0$, and $m_i/n_i\geq\eta$, together with the structural validity checks, but does not require $v_i^R=1$ or positive retrieved/retained answer overlap. Eligible trajectories use the same $\rho$ and weight transformation in Eq.~\ref{eq:redistribution}; local-credit validity is unchanged. This compares structurally eligible answer-region redistribution with CREDO's evidence-conditioned configuration.

For the main 8B comparisons, GRPO, CREDO, CWO, Random, and No-gate use monitoring-set accuracy to select among their available saved checkpoints, followed by evaluation under identical settings. All methods share the selection criterion, while candidate sets depend on checkpoint availability. In Stage~1, checkpoints for CREDO, CWO, Random, and No-gate were saved every 10 steps; in Stage~2, checkpoints were saved every 100 steps. All methods initialize Stage~2 from their own Stage~1 step-80 checkpoint. The final 500-question evaluation scores are not used for model selection in these main comparisons.

\noindent\textbf{Evaluation protocol.} We adopt normalized substring matching to retain compatibility with Temp-R1's public evaluation implementation~\citep{gong2026tempr1}. We verified the scoring behavior at commit \href{https://github.com/zjukg/Temp-R1/tree/44ab58fc81fa40d2279a172da25674075404b5e2}{\texttt{44ab58f}}, with validation format and retrieval bonuses set to zero. Its evaluation entry point routes MultiTQ and TimelineKGQA to \texttt{qa\_em\_format.compute\_score\_em}; despite the function name, the underlying \texttt{em\_check} tests substring containment rather than string equality. Extracted answers are lowercased, stripped of punctuation and English articles, and whitespace-normalized. A prediction receives one point if it contains any normalized acceptable answer; missing answers and execution failures receive zero. The earlier Search-R1 code also provides a normalized substring scorer~\citep{jin2025searchr1}, alongside an exact-match scorer. Our choice follows the scoring branch used by the direct temporal-agent baseline, rather than assuming that all prior implementations use identical metrics. The same rule is applied to all in-house comparisons. This establishes protocol consistency while retaining the metric's answer-containment interpretation; additional candidates can still receive credit.

The main MultiTQ comparison uses one evaluation configuration for CREDO, CWO, Random, and No-gate, giving Overall scores of 82.4\% (412/500), 80.2\% (401/500), 79.0\% (395/500), and 78.4\% (392/500), respectively. Evaluation uses seed~42, temperature~1, top-$p$~0.95, generation/container limits of 5,000/6,200, Information limit~2,400, retrieval top-$k$~30, and up to three searches plus a final round.

\noindent\textbf{MultiTQ baseline sources.} Our in-house outcome-only GRPO baseline shares CREDO's SFT initialization, training data, terminal reward, answer-extraction rules, and base training and evaluation settings; local credit and outcome redistribution are disabled. It scores 385/500 (77.0\%) and provides the GRPO Stage~2 endpoint in Figure~\ref{fig:credo-diagnostics}(a), the predictions for Tables~\ref{tab:app-types}--\ref{tab:app-paired}, and the frozen-model diagnostics. Table~\ref{tab:multitq-main} and Figure~\ref{fig:ablation} retain the published Temp-R1 reference of 78.0\%~\citep{gong2026tempr1}. The literature comparison and our matched reproduction therefore use distinct result sources: CREDO's Overall improvements are 4.4 and 5.4 percentage points, respectively. The final-stage GRPO, CWO, and CREDO scores use the common evaluation configuration. Earlier intermediate-stage values in Figure~\ref{fig:credo-diagnostics}(a) remain the reported records; the curve does not constitute a controlled re-evaluation of every intermediate checkpoint.

\noindent\textbf{TimelineKGQA reproduction.} The public materials available for our experiments did not provide the complete TimelineKGQA training pipeline needed to reproduce the published Temp-R1 models. We therefore train Temp-R1 and CREDO in our implementation with identical base configurations, including initialization, training data, optimizer settings, training budget, and retrieval environment. Evaluation follows the public normalized substring protocol, with identical inputs, decoding settings, token limits, interaction rules, and extraction routines. This provides a controlled comparison within our training setup. The reproduced Temp-R1 scores (50.8\% Cron, 30.9\% ICEWS) differ from the published 70.5\%/64.2\%~\citep{gong2026tempr1}; we do not claim numerical reproduction of those models or superiority over their reported scores.

Each model is evaluated on 800 Cron and 800 ICEWS rows, with correct counts of 406 versus 418 and 247 versus 291. Cron-trained checkpoints transfer directly to ICEWS without further training. Simple/Medium/Complex counts are 250/245/305 for Cron and 351/276/173 for ICEWS. The sets contain 797 and 793 unique question/answer keys; duplicate rows remain in the averages.

PoK is not included in our TimelineKGQA reproduction because a complete public implementation was unavailable to us. Its published Cron/ICEWS scores of 65.1\%/60.2\%, reported in Temp-R1, remain external reference results. Other adopted rows in Table~\ref{tab:timeline-main} provide context; the controlled comparison is reproduced Temp-R1 versus CREDO, rather than an exhaustive ranking of published systems.

Table~\ref{tab:sensitivity} uses Llama-3.2-3B-Instruct. All 3B runs use 80 training steps with a training batch size of 32; the corresponding 8B Stage~1 runs use 80 steps with a batch size of 60. Within the 3B study, all settings share the same SFT initialization, training data, base training configuration, and evaluation protocol. Only $\lambda_{\mathrm{local}}$ or $\rho$ is varied relative to the default. Evaluation uses the same settings and official 500-question MultiTQ set as the main 8B experiments, comprising 340 Single and 160 Multiple questions. The table reports only Multiple accuracy.

\subsection{Supplementary Results}
\label{app:paired}
\label{app:all-wrong}
\label{app:evidence-tests}
\label{app:sensitivity}

\noindent\textbf{Formal component comparisons.} CREDO's observed Overall gain is 2.2 percentage points over CWO and 3.4 points over the selected Random model. The CREDO/CWO Multiple comparison is 105/160 versus 99/160, a 3.75-point difference. CREDO, CWO, Random, and No-gate use the same best-monitoring-checkpoint criterion and final evaluation settings. No-gate obtains 78.4\% Overall and 57.5\% Multiple (92/160); CREDO's point estimates are higher by 4.0 and 8.13 percentage points, respectively. The component figure also includes Random's reported Multiple accuracy of 57.5\% and the literature GRPO reference (78.0\% Overall, 55.0\% Multiple) from Temp-R1. The latter is a reported baseline rather than a new independent run.

\noindent\textbf{Question-type results.} Table~\ref{tab:app-types} gives the complete supplementary breakdown. Each model has 500 unique question identifiers with matching question/answer hashes. Missing extracted answers number 3, 3, and 6 for GRPO, CWO, and CREDO and remain in the denominator. CREDO's gains over GRPO cover all three Multiple categories, while Single-category gains are not uniform.

\begin{table}[!htbp]
\centering\small
\caption{Paired MultiTQ results: correct answers (accuracy, \%). CREDO and CWO also provide the main results; GRPO is our matched in-house baseline. Aggregate rows overlap the individual categories.}
\label{tab:app-types}
\setlength{\tabcolsep}{4pt}
\begin{tabular*}{\linewidth}{@{\extracolsep{\fill}}lrrrr@{}}
\toprule
Category & $N$ & GRPO & CWO & CREDO\\
\midrule
Equal & 174 & 156 (89.66) & 151 (86.78) & 152 (87.36)\\
Before/After & 94 & 81 (86.17) & 89 (94.68) & 87 (92.55)\\
First/Last & 72 & 65 (90.28) & 62 (86.11) & 68 (94.44)\\
Equal Multi & 33 & 22 (66.67) & 28 (84.85) & 28 (84.85)\\
Before Last & 69 & 34 (49.28) & 38 (55.07) & 41 (59.42)\\
After First & 58 & 27 (46.55) & 33 (56.90) & 36 (62.07)\\
\midrule
Single & 340 & 302 (88.82) & 302 (88.82) & 307 (90.29)\\
Multiple & 160 & 83 (51.88) & 99 (61.88) & 105 (65.63)\\
Entity & 371 & 263 (70.89) & 282 (76.01) & 286 (77.09)\\
Time & 129 & 122 (94.57) & 119 (92.25) & 126 (97.67)\\
Overall & 500 & 385 (77.00) & 401 (80.20) & 412 (82.40)\\
\bottomrule
\end{tabular*}
\end{table}

\noindent\textbf{Paired comparisons.} Table~\ref{tab:app-paired} reports corrections, regressions, and net differences on the same inputs. Intervals use 100,000 paired question bootstrap samples, seed~20260925, and percentile endpoints at 2.5\% and 97.5\%, without multiple-comparison correction. Against the matched in-house GRPO checkpoint, corrections exceed regressions by 27 questions Overall and 22 on Multiple, with both paired intervals above zero. The CWO intervals include zero and quantify uncertainty in the main CREDO--CWO difference. These intervals are conditional on the selected checkpoints and quantify question-sampling uncertainty, not training-seed variation or checkpoint-selection uncertainty.

\begin{table}[!htbp]
\centering\small
\caption{Paired supplementary comparisons. ``Corrected'' means only CREDO is correct; ``Regressed'' means only the comparator is correct. Differences and intervals are in percentage points.}
\label{tab:app-paired}
\setlength{\tabcolsep}{3pt}
\begin{tabular*}{\linewidth}{@{\extracolsep{\fill}}lrrrrr@{}}
\toprule
Comparator / subset & $N$ & Corrected & Regressed & Difference & 95\% CI\\
\midrule
GRPO / Overall & 500 & 50 & 23 & $+5.40$ & $[2.20,8.80]$\\
CWO / Overall & 500 & 34 & 23 & $+2.20$ & $[-0.80,5.20]$\\
GRPO / Multiple & 160 & 33 & 11 & $+13.75$ & $[6.25,21.88]$\\
CWO / Multiple & 160 & 19 & 13 & $+3.75$ & $[-3.13,10.63]$\\
\bottomrule
\end{tabular*}
\end{table}

\noindent\textbf{All-wrong groups.} Across 80 Stage~1 updates, 4,800 question-group exposures yield 24,000 trajectories ($K=5$). Of these groups, 1,473 are all-wrong (30.69\%); 564 have active local credit (11.75\% of all groups; 38.29\% of all-wrong groups). Both curves use all groups as the denominator. Rewards are binary substring correctness, with format/retrieval bonuses disabled and KL excluded, so all-wrong groups have zero outcome advantages. Within active all-wrong groups, retrieval, selection, and reasoning teachers occur in 351, 350, and 59 groups, respectively (overlapping counts). These observations establish local-signal availability at all three stages, not gradient magnitude or an isolated performance contribution.

\noindent\textbf{Single-question distractors.} Among the fixed 500 inputs, 84 questions have parseable explicit dates, 72 pass rule-based KG answer checks and finite evidence construction, and 67 admit matched distractor conditions. All 67 are Single questions with entity answers: 60 Equal and 7 Before/After, with day/month/year counts 51/13/3. They are not an independently human-labeled evidence-sufficiency set.

The prompt instructs the model to answer from supplied evidence without tools and end with an Answer block. The user input contains the question and numbered Information lines. Retrieval is disabled; decoding uses temperature~1, top-$p$~0.95, generation limit~1,024, context limit~11,200, and seeds~42/43/44 with question-hash offsets. The first complete Answer block stops generation. Over-budget inputs are rejected rather than truncating support. Generic and temporal distractors are matched in count, length, format, and insertion position. The former are unrelated facts; the latter are topically related but violate the question's temporal condition. Supporting facts remain fixed for $k\in\{0,2,4,8\}$; an additional condition contains unrelated evidence only.

\begin{table}[!htbp]
\centering\small
\caption{Answer containment under fixed evidence. Entries are correct generations out of 201 (substring accuracy, \%), from 67 questions and three decoding seeds.}
\label{tab:app-evidence}
\begin{tabular*}{\linewidth}{@{\extracolsep{\fill}}lrrr@{}}
\toprule
Evidence condition & GRPO & CWO & CREDO\\
\midrule
Supporting facts only & 178 (88.56) & 195 (97.01) & 194 (96.52)\\
Generic, $k=2$ & 164 (81.59) & 184 (91.54) & 196 (97.51)\\
Generic, $k=4$ & 170 (84.58) & 184 (91.54) & 198 (98.51)\\
Generic, $k=8$ & 154 (76.62) & 181 (90.05) & 196 (97.51)\\
Temporal, $k=2$ & 157 (78.11) & 180 (89.55) & 196 (97.51)\\
Temporal, $k=4$ & 152 (75.62) & 178 (88.56) & 197 (98.01)\\
Temporal, $k=8$ & 142 (70.65) & 183 (91.04) & 196 (97.51)\\
Unrelated evidence only & 4 (1.99) & 3 (1.49) & 15 (7.46)\\
\bottomrule
\end{tabular*}
\end{table}

The temporal penalty subtracts temporal-condition accuracy from generic-condition accuracy at matched $k$; negative values favor temporal distractors. Table~\ref{tab:app-evidence} pools three decoding repetitions per question, with equal numbers of repetitions for every question. These are 67 questions, not 201 independent samples or training runs. This Single-question result concerns target containment with retained support; it does not directly test unique-answer selection, retrieval quality, or Multiple-question composition.

\needspace{6\baselineskip}
\noindent\textbf{Multiple-question distractors.} We additionally evaluate 60 Multiple questions, with 20 from each official type: Equal Multi, Before Last, and After First. The 8B models are GRPO at step~200 and CWO/CREDO at step~400. Each model answers every question under three decoding seeds in each condition, yielding 180 generations per model and condition. Complete supporting facts and sorting candidates are retained. Generic and temporal conditions each add four facts; inputs are identical across models, retrieval is disabled, and no input is truncated. Scoring follows the same official normalized substring protocol.

\begin{table}[!htbp]
\centering\small
\caption{Multiple-question answer containment with complete support retained. Entries are correct generations out of 180 (substring accuracy, \%), from 60 questions and three decoding seeds. Bold indicates the highest accuracy in each condition.}
\label{tab:app-multiple-distractors}
\begin{tabular*}{\linewidth}{@{\extracolsep{\fill}}lrrr@{}}
\toprule
Evidence condition & GRPO & CWO & CREDO\\
\midrule
Complete evidence & 60 (33.33) & 114 (63.33) & \textbf{125 (69.44)}\\
Generic, $k=4$ & 58 (32.22) & 96 (53.33) & \textbf{119 (66.11)}\\
Temporal, $k=4$ & 62 (34.44) & 97 (53.89) & \textbf{100 (55.56)}\\
\bottomrule
\end{tabular*}
\end{table}

CREDO achieves the highest absolute accuracy in every condition (Table~\ref{tab:app-multiple-distractors}). Under temporal distraction, its advantage over CWO narrows to three correct generations (1.67 percentage points). Relative to complete evidence, CREDO loses 13.89 percentage points and CWO loses 9.44; relative to generic distraction, CREDO loses 10.56 points. Thus, the advantage in absolute accuracy coexists with greater sensitivity to temporal distraction in this Multiple setting. These results extend the evaluation to compositional question types, but do not establish uniformly stronger temporal robustness or isolate correct reasoning from answer containment. The 180 generations represent repeated decoding on 60 questions, not independent questions or training runs.

\needspace{9\baselineskip}
\noindent\textbf{Hyperparameter sensitivity.} Starting from the default $(\lambda_{\mathrm{local}},\rho)=(0.03,0.04)$, we vary $\lambda_{\mathrm{local}}\in\{0.01,0.03,0.05\}$ with $\rho$ fixed, and vary $\rho\in\{0.02,0.04,0.06\}$ with $\lambda_{\mathrm{local}}$ fixed. All runs share the same 3B backbone, SFT initialization, training data, base training configuration, and evaluation protocol. Every row follows the same reporting rule: evaluate the available candidate checkpoints on the official 500-question MultiTQ set using normalized substring matching, and report the highest accuracy on the same 160 Multiple questions. This rule applies to the default configuration and all coefficient variants. The table therefore compares observed peak performance across parameter settings; unlike the main 8B comparisons, checkpoint selection in this exploratory analysis uses the reported evaluation subset. The results do not estimate performance under independent monitoring-based selection or variability across training seeds.

\CREDOclearwrap

\subsection{Discussion and Limitations}
\label{app:limitations}

\noindent\textbf{Evidence proxies.} Answer overlap does not test semantic sufficiency. With answer tokens preserved, all 40 rule-checked Single examples retain positive coverage in each of the original, wrong-time, and wrong-relation conditions. This diagnoses coverage alone, not the full provenance/eligibility pipeline. Natural-trajectory gate precision and recall remain unmeasured; aggregated masks provide stage-level, not per-action, credit.

\noindent\textbf{Mechanism and stability.} Component comparisons do not isolate all-wrong updates, component interactions, or individual local-credit ingredients. Best-checkpoint comparisons describe the selection protocol rather than fixed-update effects. The training trace and 3B study use single runs; decoding repetitions and paired question intervals do not measure training-seed stability.

\end{document}

%% file: math_commands.tex
\usepackage{amsmath,amsfonts,bm}

\def\eqref#1{equation~\ref{#1}}

\def\1{\bm{1}}

\DeclareMathAlphabet{\mathsfit}{\encodingdefault}{\sfdefault}{m}{sl}
\SetMathAlphabet{\mathsfit}{bold}{\encodingdefault}{\sfdefault}{bx}{n}

